\documentclass[runningheads]{llncs}
\usepackage[T1]{fontenc}
\usepackage{graphicx}
\usepackage{booktabs}
\usepackage[misc]{ifsym}
\newcommand{\corr}{(\Letter)}
\usepackage{mwe}
\usepackage{comment}
\usepackage{amsmath} 
\usepackage{nicefrac}
\usepackage{microtype}
\usepackage{float}
\usepackage{subcaption}
\PassOptionsToPackage{hyphens}{url}\usepackage{hyperref}
\usepackage{cleveref}
\usepackage{xurl}
\usepackage{booktabs}
\usepackage{multirow}
\usepackage{makecell}
\usepackage{tabularx}
\usepackage[separate-uncertainty=true]{siunitx}
\usepackage[subtle]{savetrees}

\begin{document}
  
%\title{From Instructions to Assistance: The Manual-to-Action Dataset for the Evaluation of Multimodal Language Models}
\title{From Instructions to Assistance: a Dataset Aligning Instruction Manuals with Assembly Videos for Evaluating Multimodal LLMs}
\titlerunning{From Instructions to Assistance}
% If the full title of your paper is short enough to also fit in the running head, you can omit the abbreviated paper title here. You can check as follows: if you comment out the \titlerunning line, something will appear in the header of all odd-numbered pages of your PDF from page 3 onward. This something is either the full title (in which case all is well), or the error message "Title Suppressed Due to Excessive Length". If this error message appears, you're going to want to provide an abbreviated title within the \titlerunning command, because if you won't do it, Springer will do it for you.

%N.B.: Author information (both in the \author{} and \authorrunning{} command) should only be present in the Camera-Ready Version of your paper. The version that you initially submit for review, ought to be double-blind. So, when initially submitting your paper, use:
% \author{Author information scrubbed for double-blind reviewing}

%\author{Author information scrubbed for double-blind reviewing}
\author{
     Federico Toschi\inst{1} \corr \and 
     Nicolò Brunello\inst{1} \and 
     Andrea Sassella\inst{1} \and 
     Vincenzo Scotti\inst{2} \and 
     Mark James Carman\inst{1}
 }

% You may leave out the orcidID information, if you want to.
% Use \corr to indicate the corresponding author. Note the spacing around the \corr command. Only one author can be the corresponding author.

%N.B.: comment out the \authorrunning{} command for the double-blind version of your paper submitted for review. Later, if your paper is accepted, use the command for the Camera-Ready Version.
\authorrunning{Toschi et al.}
% First names are abbreviated in the running head.
% If there is one author, write 'A.L. Benjamin'.
% If there are two authors, write 'A.L. Benjamin and C.C. Broadus Jr.'
% If there are more than two authors, '[...] et al.' is used.

\institute{
DEIB, Politecnico di Milano, Via Ponzio 34/5, 20133, Milano (MI), Italy \\
     \email{federico.toschi@polimi.it} \qquad \email{nicolo.brunello@polimi.it} \\
     \email{andrea.sassella@polimi.it} \qquad \email{mark.carman@polimi.it}
 \and
 KASTEL, Karlsruhe Institute of Technology, Am Fasanengarten, 576131, Karlsruhe, Germany \\
 \email{vincenzo.scotti@kit.edu}
}

\maketitle              % typeset the header of the contribution

\begin{abstract}
The recent advancements introduced by Large Language Models (LLMs) have transformed how Artificial Intelligence (AI) can support complex, real-world tasks, pushing research outside the text boundaries towards multi-modal contexts and leading to Multimodal Large Language Models (MLMs). 
Given the current adoption of LLM-based assistants in solving technical --or domain-specific-- problems, the natural continuation of this trend is to extend the input domains of these assistants exploiting MLMs.
Ideally, these MLMs should be used as real-time assistants in procedural tasks, hopefully integrating a view of the environment where the user being assisted is, or even better sharing the same point of view via Virtual Reality (VR) or Augmented Reality (AR) supports, to reason over the same scenario the user is experiencing.
With this work, we aim at evaluating the quality of currently openly available MLMs to provide this kind of assistance on technical tasks.
To this end, we annotated a data set of furniture assembly with step-by-step labels and manual references: the Manual-to-Action Dataset (M2AD).
We used this dataset to assess (1) to which extent the reasoning abilities of MLMs can be used to reduce the need for detailed labelling, allowing for more efficient, cost-effective annotation practices, (2) whether MLMs are able to track the progression of assembly steps (3) and whether MLMs can refer correctly to the instruction manual pages.
Our results showed that while some models understand procedural sequences, their performance is limited by architectural and hardware constraints, highlighting the need for multi-image and interleaved text-image reasoning.
\end{abstract}

\section{Introduction}

\emph{Large Language Models} (LLMs) now define the cutting edge of \emph{Artificial Intelligence} (AI) innovation.
Their advancements in text processing are remarkable~\cite{DBLP:journals/corr/abs-2402-06196} and since the last couple of years, their capabilities are not even limited to text as they have been extended into \emph{Multimodal Language Models} (MLMs)~\cite{DBLP:journals/corr/abs-2306-13549}.
These MLMs integrate multiple input modalities besides text, including images, videos and audio. 
This shift to multiple input modalities provide with a broader understanding of complex, real-world tasks with respect to base LLMs. 
However, as these models grow in capability, evaluating them becomes increasingly challenging due to the added intricacies of multimodal tasks. Existing benchmarks largely focus on isolated skills, such as image-text alignment, and fall short of capturing the depth required for more complex tasks, particularly in scenarios that demand sequential understanding, real-time interaction, or procedural reasoning~\cite{bai2024surveymultimodal}. 

One emerging application of LLMs has been that of assistance in technical, or --in general-- domain-specific, problem solving~\cite{DBLP:conf/pkdd/ScottiC24,fornasiere-etal-2024-medical}.
Normally, in these scenarios we would have a technician, the \emph{expert}, helping a novice or a customer, \emph{the user}, solve some task.
What current technology allows is to have the LLM assist the expert in supporting the user generating suggestions in real-time.
Thus, the natural extension of this problem would be to adopt MLMs and extend the set of input modalities.
For the specific case of technical assistance, we would like to focus on the situations where the model is provided with additional visual input of the environment, whether it is a \emph{exo-centric} (i.e., third person) view from an external camera or an \emph{ego-centric} (i.e., first person) view, obtained for example via \emph{Virtual Reality} (VR) or \emph{Augmented Reality} (AR).
From the video, the MLM can capture a view of the environment and the problem to solve, hopefully sharing this view with the user as is becoming common in many industrial settings~\cite{DBLP:conf/ism2/BarattaCG023,DBLP:journals/thri/RobinsonTCKC23}.

In these technical assistance scenarios, MLMs are expected to process text and video to help users navigate step-by-step tasks, predict actions, and even detect potential errors~\cite{Plizzari2024outlookego}.
Although a few benchmarks, like \emph{HoloAssist}~\cite{wang2023holoassist} or \emph{ENIGMA-51}~\cite{Ragusa2024enigma51}, try to address the problems of action prediction or error detection, they often lack a deeper evaluation needed assess models goodness in complex, multi-step, tasks like furniture assembly.
In this sense, to our knowledge, only the \emph{IKEA Manuals At Work}~\cite{liu2024ikea_manuals_at_work} includes the appropriate data, but it still focuses on the same simpler tasks of previous datasets and as them comes with a set of complex annotations we believe newer end-to-end MLMs shouldn't need.

To address these limitations we propose a benchmark designed for multi-step multi-modal reasoning: the \emph{Manual-to-Action Dataset} (M2AD).
M2AD features video clips showing, from different point of views, the assembly procedure of pieces of furniture from IKEA and related instruction manuals, annotated with step alignment labels to map between the video and the manual. 
We designed this data set to address specifically the use of MLMs in multi-step, real-world assistance, testing their reliability in procedural environments that demand a fine-grained understanding of both visual and textual cues.
For completeness, we assessed baselines on M2AD using openly available MLMs to understand the level of state-of-the-art models that can be run on consumer-level hardware in (1) tracking the completion status of an assembly process, (2) detecting whether a step from the instruction manual corresponds to that of a video segment and (3) identifying the current step being executed.

We divide the rest of this paper into the following sections.
In \Cref{sec:relatedwork}, we discuss on recent advances about MLMs and about available datasets for evaluating visual assistants.
In \Cref{sec:m2ad}, we characterise our M2AD dataset, explaining motivations and annotation process.
In \Cref{sec:experiments} we report about baseline experiments we conducted on our dataset with openly available multimodal models.
In \Cref{sec:conclusion} we summarise our contribution and suggest possible future extensions.

\section{Related Work}
\label{sec:relatedwork}

In this section we present and discuss the related works about MLMs (\Cref{sec:mlms}) and existing datasets for multimodal technical assistance (\Cref{sec:relateddata}).

\subsection{Multimodal Language Models}
\label{sec:mlms}

MLMs extend LLMs integrating multiple data modalities (images, audio, and video), enabling more flexible AI applications. 
These models typically consist of three modules: a modality encoder, a language model, and a modality interface. 

The modality encoder, often derived from computer vision research, converts non-textual inputs into embeddings that the language model can process. Common encoding techniques include \emph{Convolutional Neural Networks} (CNNs) like \emph{ResNet}~\cite{DBLP:conf/cvpr/HeZRS16} and \emph{Vision Transformers} (ViT)~\cite{DBLP:conf/iclr/DosovitskiyB0WZ21}, which apply transformer models to visual tasks by splitting images into patches.
% Vision Language Models (VLMs), a subset of MLLMs, are trained using methods like Language Image Contrastive Learning and Self-Supervised Learning to align visual and textual embeddings, making them more robust and effective for multimodal tasks.
Often the embedding spaces are aligned with contrastive learning techniques like CLIP~\cite{DBLP:conf/icml/RadfordKHRGASAM21}.

The modality interface is crucial for bridging the gap between the embeddings generated by the modality encoder and the language model. 
There are three primary techniques for implementing this interface: \emph{token-level fusion}, \emph{feature-level fusion}, and \emph{hybrid approaches}. 
Token-level fusion involves transforming encoder outputs into tokens that can be concatenated with text tokens and input into the LLM~\cite{liu2023llava,liu2023improvedllava,liu2024llavanext,li2022blip,zhang-etal-2023-video,dai2023instructblip}. 
Feature-level fusion enables deeper interactions between modalities by incorporating cross-attention layers within the LLM, allowing the model to attend to both textual and visual features~\cite{alayrac2022flamingo}. 
Hybrid approaches combine these methods~\cite{zhang2023internlmxcomposervisionlanguagelargemodel,dong2024internlmxcomposer2masteringfreeformtextimage}, often using additional modules like perceiver samplers to reduce dimensionality and fine-tuning techniques like \emph{Low-Rank Adaptation} (LoRA)~\cite{DBLP:conf/iclr/HuSWALWWC22} to enhance the model's ability to utilize visual information effectively.

Prominent and best performing MLMs include commercial models like \emph{GPT}~\cite{DBLP:journals/corr/abs-2303-08774,DBLP:journals/corr/abs-2410-21276} and \emph{Gemini}~\cite{DBLP:journals/corr/abs-2312-11805,DBLP:journals/corr/abs-2403-05530} which are closed-access and accessible only via external APIs.
However, there exist openly accessible alternatives, which include \emph{LLaMa 3.2 Vision}~\cite{meta2024llama32}(LLaMa 3.2 was not accessible in EU when we wrote this paper), \emph{Fuyu}~\cite{fuyu-8b} or other models obtained fine-tuning existing LLMs and MLMs.
Models based on LLMs include \emph{LLaVa}~\cite{zhang2024llava_video,li2024llava_onevision} (based either on \emph{LLaMa}~\cite{DBLP:journals/corr/abs-2307-09288,DBLP:journals/corr/abs-2407-21783} or \emph{Qwen 2}~\cite{DBLP:journals/corr/abs-2407-10671}), \emph{MolMo}~\cite{deitke2024molmopixmoopenweights} (again based on \emph{Qwen 2}) and \emph{Ovis}~\cite{lu2024ovis} (based on \emph{Gemma 2}~\cite{DBLP:journals/corr/abs-2408-00118}) while those obtained extending MLMs --usually to support visual instruction following include \emph{MFuyu}~\cite{jiang2024mantis} (derived from the MLM \emph{Fuyu}).
In this work, we will focus on open access models that can run on consumer-level hardware, respectively, for reproducibility and to show that it is possible to maintain data confidentiality, which can be a concern in industrial settings, without needing excessive computational resources.
Overall, MLMs represent a significant advancement in AI, enabling more comprehensive and context-aware understanding of multimodal data.

\subsection{Multimodal Assistance Datasets}
\label{sec:relateddata}

The landscape of multimodal datasets and benchmarks for procedural tasks is rich and varied, encompassing a wide range of settings, annotations, and tasks. 
These datasets are designed to evaluate the capabilities of models in understanding and executing complex, multi-step activities, often requiring the integration of multiple modalities such as video, audio, gaze tracking, and depth information. 
As a result, the tasks related to the problem of video understanding span across a lot of domains and topics: (1) cooking and household tasks, (2) industrial and assembly tasks, (3) technical assistance and human-object interaction, (4) multi-domain and general tasks and (5) specialized procedural tasks.

The tasks evaluated by these datasets are equally diverse, ranging from action recognition and error detection to assembly plan generation and next step planning. 
Similarly, the point of view represents another critical characteristic, with datasets adopting either an egocentric perspective or an exocentric perspective.

Among the datasets related to cooking and households domain, we have \emph{EGTEA Gaze+}~\cite{li2018egtea_gaze}, which focuses on cooking activities and provides frame-level action annotations along with video and gaze data. Similarly, \emph{EPIC-KITCHENS}~\cite{damen2018epic_kitchens} extends its scope to include both cooking and house chores, offering action segments and incorporating video and audio modalities. 
\emph{CaptainCook4D}~\cite{peddi2023captaincook4d} focuses on cooking as well, providing action labels and error descriptions, and utilizing video and audio data. 
\emph{EgoPer}~\cite{lee2024egoper} provides action labels with video, depth, gaze, and audio data, and \emph{EgoPlan-Bench}~\cite{chen2024egoplanbenchbenchmarkingmultimodallarge} provides task goals and video data, both for cooking.

Concerning industrial and assembly tasks, \emph{Assembly101}~\cite{sener2022assembly101} targets industrial settings without instructions, providing action and error labels, and relying solely on video data. \emph{MECCANO}~\cite{ragusa2023meccano} \emph{ENIGMA-51}~\cite{Ragusa2024enigma51} also focus on industrial settings but include instructions, the former offering temporal annotations for actions interactions, utilizing video, depth, and gaze data, and the latter offering action labels and questions from workers, and utilizing video and audio data. \emph{IndustReal}~\cite{schoonbeek2024industreal} provides action labels and assembly state annotations, and utilizing video, depth, and gaze data. 
\emph{IKEA Manuals at Work}~\cite{liu2024ikea_manuals_at_work} focuses on assembly tasks, providing assembly state annotations and utilizing video and masks. 

About technical assistance, {HoloAssist}~\cite{wang2023holoassist} is particularly relevant for the topic, offering transcriptions of conversations with purpose labels and action labels, and incorporating video, gaze, depth, and audio data. 
\emph{HOI4D}~\cite{liu2022hoi4d} delves into human-object interactions, using fine-grained action labels with video and depth information. 
\emph{EgoExoLearn}~\cite{huang2024egoexolearn} spans cooking and lab settings, offering action labels and skill level annotations with video, gaze, and audio data. 

Multi-domain and general tasks are covered in \emph{Ego4D}~\cite{grauman2022ego4d}, which stands out for its multi-domain approach, featuring clip captions and video summaries, and incorporating video, gaze, and audio data. 
Multi-domain examples are also covered in \emph{EgoObjects}~\cite{zhu2023ego_objects}, which provides object class annotations with video data, and in \emph{EgoOops}~\cite{haneji2024egooopsdatasetmistakeaction}, which focuses on electrical circuits, chemistry, and crafts, providing procedure step annotations and using video and audio data. 

From the specialised procedural tasks group we have only \emph{EPIC-TENT}~\cite{jang2019epictent}.
This dataset focuses on tent building, providing task labels with video and gaze data. 

However, we believe that these datasets, particularly in the context of assembly tasks and technical assistance tasks (which require a deeper understanding of context and cumulative states over time), are limited. 
This limitation is mainly due to the reliance on the heavy and atomic level of the annotations about the tasks evaluated, which do not fully align with the complexities of real-world technical assistance scenarios.
% With our dataset M2AD we try to propose an alternative benchmark tackling this problem. 
With M2AD we try to address this problem.

\section{M2AD dataset}
\label{sec:m2ad}

In this section, we introduce M2AD, its motivations and its annotation process.

% In this section, we provide an overview of M2AD (\Cref{sec:overview}), we discuss its motivations comparing it with existing benchmarks (\Cref{sec:motivations}) and we detail the annotation process we followed to create it from the raw data (\Cref{sec:annotationprocess}).

\subsection{Overview}
\label{sec:overview}

The M2AD dataset offers a rich and diverse collection of annotated video-manual pairs, capturing the process of furniture assembly in realistic contexts. 
We designed M2AD to be an evaluation benchmark of MLMs used as multimodal technical assistants in the use case of furniture assembly.
The emphasis on aligning video content with instructional steps, combined with detailed temporal and behavioral analyses, makes M2AD a valuable resource for evaluating and advancing MLMs in this context.
Unlike datasets constructed in controlled environments, M2AD is derived from publicly available YouTube videos, which were carefully annotated and processed to ensure a one-to-one correspondence between video clips and an the corresponding sections in the instruction manuals scraped from the IKEA website. 
This approach ensures that the dataset reflects real-world assembly scenarios, capturing the inherent variability in human-driven tasks.

The dataset encompasses over 50 unique furniture pieces, representing a diverse range of assembly challenges and instructional content. 
Each video in the dataset is annotated to align the assembly procedures depicted in the video with the corresponding steps in the instruction manual. Specifically, annotations are linked to individual video segments that correspond to the execution of a single assembly step. 
These annotations include the start and end timestamps of the step, the step number as indicated in the manual, and the page number of the manual where the step is described. 
These annotations provide a ground truth for evaluating model's accuracy to interpret and align multimodal procedural data.

\begin{table}[!ht]
\caption{dataset summary statistics for the 53 furniture items considered.}
\label{tab:dataset_stats}
\begin{center}
% \resizebox{0.45\textwidth}{!}{
\begin{tabular}{lS[table-format=7.1]S[table-format=7.1]S[table-format=7.1]S[table-format=7.1]S[table-format=7.1]}

    \toprule

    & {\textbf{Total}} & {\textbf{Avg.}} & {\textbf{Std.}} & {\textbf{Min}} & {\textbf{Max}} \\
    
    \midrule
    
    \textbf{No. of Annotations} & 1228.0 \\
    \textbf{Frames per Step} & & 1365.3 \\
    \textbf{No. of annotated frames} & & 1676677.8 \\

    \midrule
    
    \textbf{No. of steps per video} & & 23.2 & 16.2 & 2.0 & 71.0 \\
    \textbf{Step duration [s]} & & 21.1 & 20.5 & 1.0 & 176.0 \\
    \textbf{Completion time [s]} & & 652.5 & 465.4 & & \\
    \textbf{Inter-step gap [s]} & & 7.3 & 18.9 \\
     
    \bottomrule

\end{tabular}  
% }
\end{center}
\end{table}

\begin{figure}[!ht]
\begin{center}
    \begin{minipage}[b]{0.45\columnwidth}
        \centering
        \includegraphics[width=\textwidth]{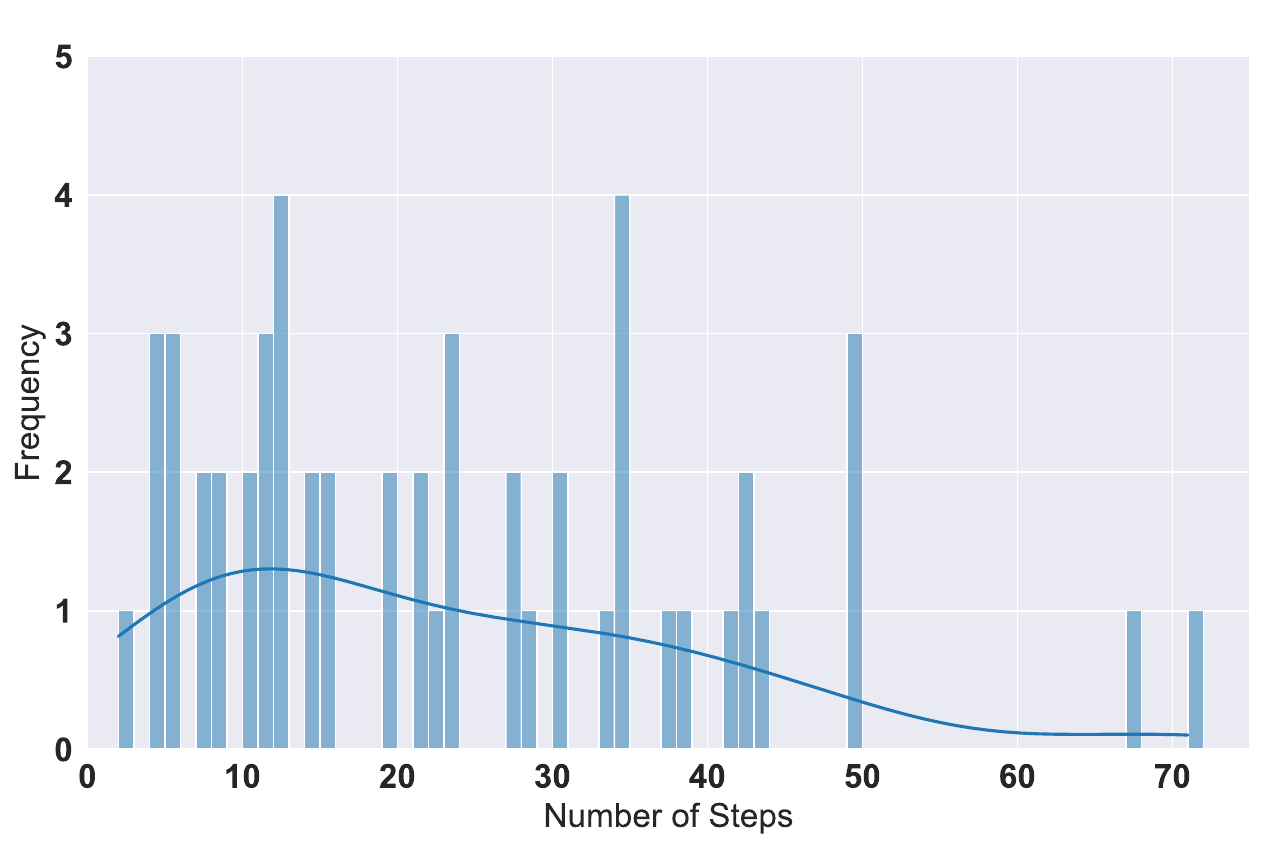}
        \caption{Distribution of the number of steps per video.}
        \label{fig:distrib_steps}
    \end{minipage}
    \hfill
    \begin{minipage}[b]{.45\columnwidth}
        \includegraphics[width=\textwidth]{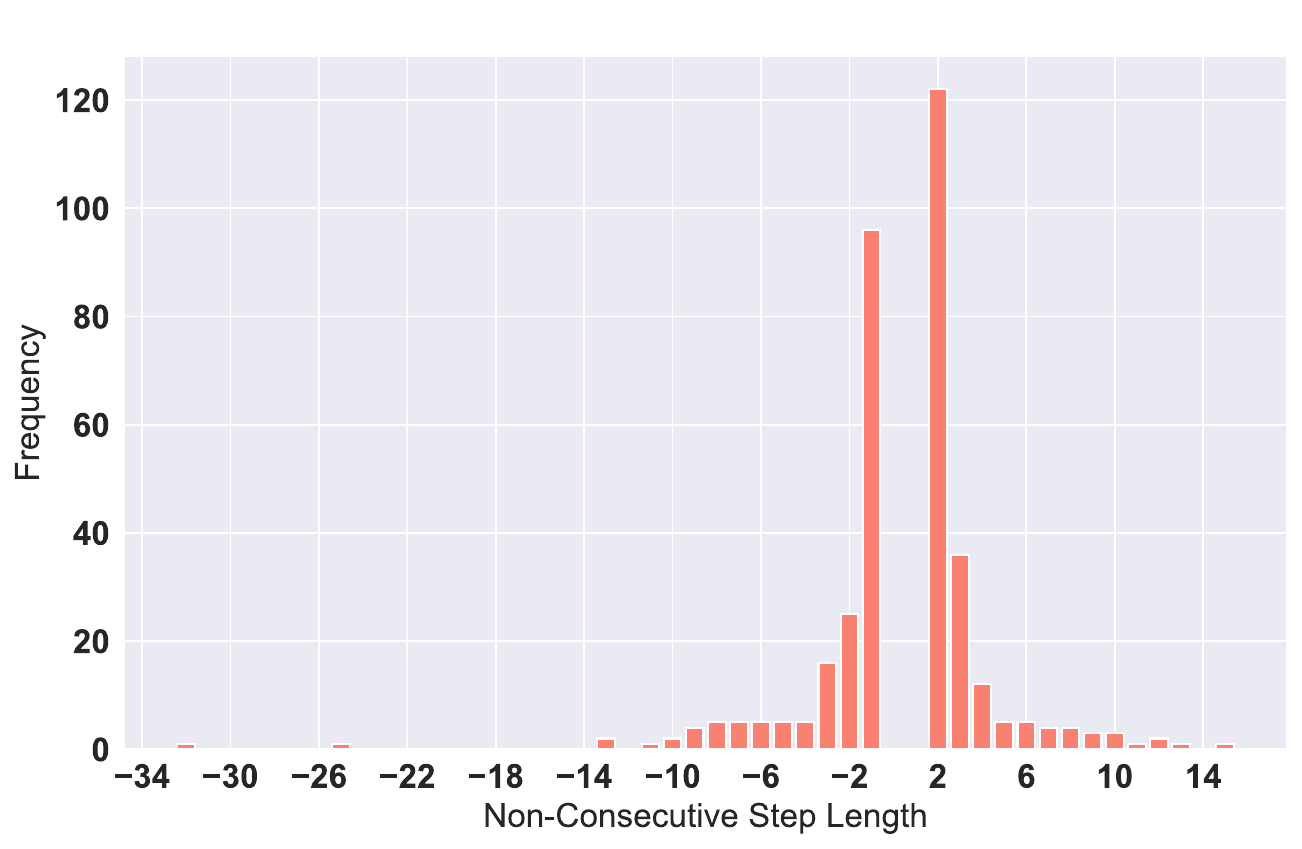}
        \caption{Distribution of non-consecutive step transition lengths.}
        \label{fig:frequency_non_cons}
    \end{minipage}
\label{fig:llm}
\end{center}
\end{figure}

We report key statistics about the dataset in \Cref{tab:dataset_stats} to quantify the dataset's scope and diversity. 
The dataset covers $53$ furniture items and comprises $1228$ annotations across all videos, with an average of $23.2$ steps per video. 
The number of steps varies significantly depending on the complexity of the furniture item, ranging from as few as $2$ steps to as many as $71$ steps per video (we visualise the distribution in \Cref{fig:distrib_steps}). 
The average duration of a single step is $21.1$ seconds, with a standard deviation of $20.$5 seconds, highlighting the variability in task complexity and user execution. 
The total completion time for assembling a piece of furniture averages $652.5$ seconds, with an average inter-step gap of $7.3$ seconds. 
These temporal metrics underscore the dataset's ability to capture the real-world pacing and interruptions that occur during manual assembly tasks.

\begin{figure}
    \centering
    \includegraphics[width=.7\linewidth]{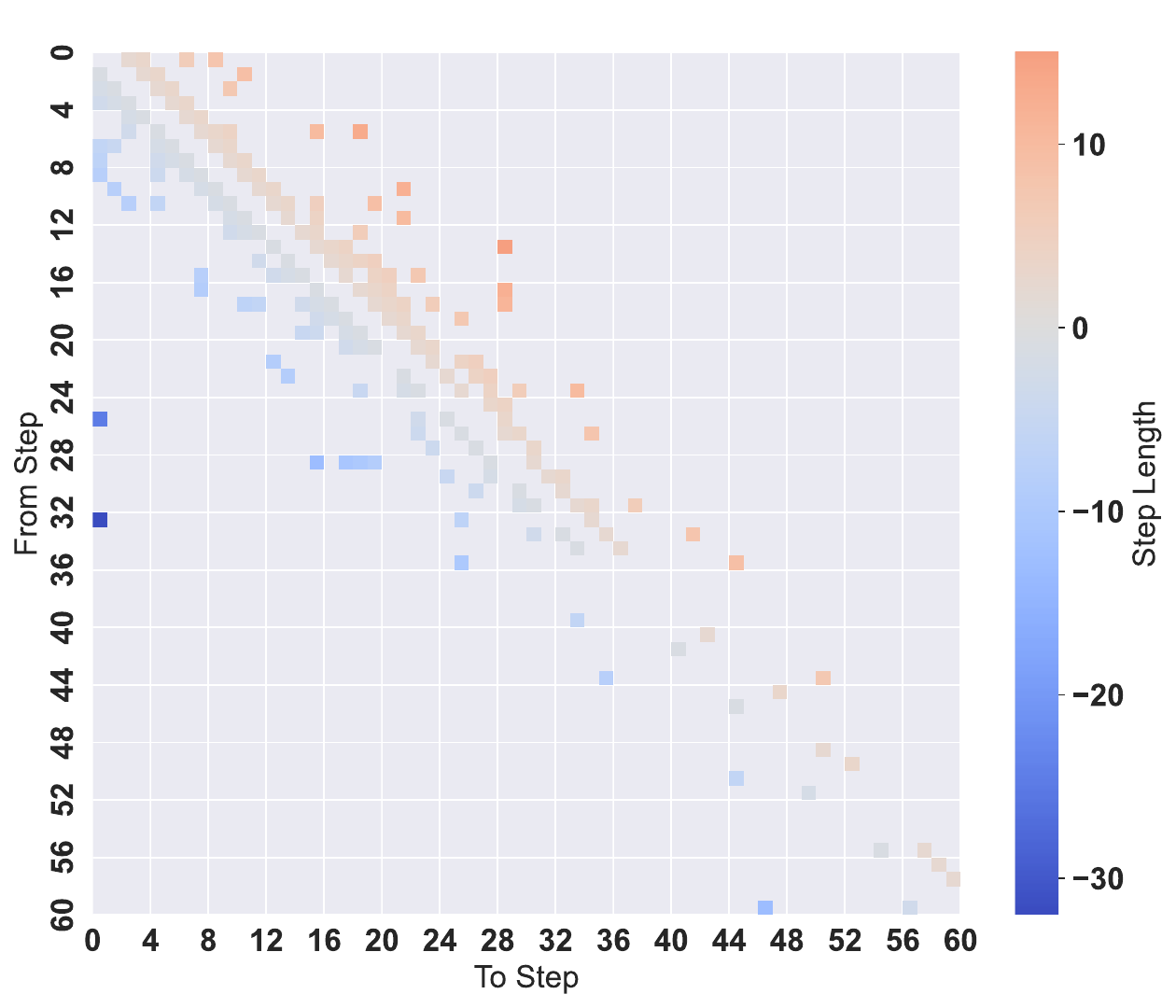}
    \caption{Heatmap showcasing the trend in non-consecutive step transition lenghts. Warm colours indicate positive transitions (skipping), and cold colours indicate negative transitions (re-visiting).}
    \label{fig:heatmap_non_cons}
\end{figure}

The dataset also includes a detailed analysis of step transitions, particularly focusing on non-consecutive transitions, which reveal patterns of user behaviour during assembly. 
By examining the lengths of forward jumps (skipping steps) and backward jumps (revisiting steps), the dataset provides insights into how users deviate from a strictly linear assembly process. 
These deviations are visualized in \Cref{fig:frequency_non_cons} through a distribution of non-consecutive step transition lengths and the heatmap in \Cref{fig:heatmap_non_cons} that illustrates the direction and magnitude of these transitions. 
The heatmap, in particular, uses warm colours to indicate forward jumps and cool colours to denote backward regressions, offering a clear visual representation of where and how users diverge from the prescribed sequence and revealing that steps in the initial phases of assembly are more interchangeable, likely because the order of tasks such as inserting dowels, screws, or locks has minimal impact on the overall outcome.

The realistic nature of the dataset is further emphasised by the observed variability in step execution and transition patterns. 
These patterns reflect the influence of user expertise, demonstrating how individuals adapt assembly procedures based on their understanding and experience. 
For instance, the distribution of non-consecutive step transitions highlights the flexibility and adaptability of human users, who may skip or revisit steps depending on their confidence or the perceived complexity of the task. 
This realism is a key strength of the dataset, as it provides a more accurate representation of how assembly tasks are performed in real-world settings, as opposed to idealized or scripted scenarios.

\subsection{Motivations}
\label{sec:motivations}

While existing datasets have contributed significantly to the evaluation of MLMs, they often fall short in evaluating all abilities required in problems like technical and assembly assistance. 
We developed M2AD  especially to address these limitations and provide a more comprehensive evaluation framework for MLMs.

Existing benchmarks, such as IKEA Manuals at Work\cite{liu2024ikea_manuals_at_work}, ENIGMA-51\cite{Ragusa2024enigma51}, and HoloAssist\cite{wang2023holoassist}, focus primarily on atomic tasks within the problem of video understanding, such as \emph{action recognition}, \emph{error detection}, and \emph{human-object interaction detection}. 
While these tasks are important, they fail at capturing the holistic nature of technical assistance, as the procedural tasks in the aforementioned datasets often evaluate models at a superficial level, focusing on recognizing individual actions or detecting errors in isolation. 
In contrast, models applied for technical assistance require a deeper level of understanding are expected to integrate multiple capabilities, such as  \emph{contextual reasoning}, \emph{intent interpretation}, and \emph{real-time guidance}, simultaneously. 
This restricted view of the tasks to solve limits the ability of existing benchmarks to assess actual MLMs performances. 

Another significant limitation of current datasets is their reliance on highly structured, labour-intensive annotations. 
Benchmarks like IKEA Manuals at Work and HoloAssist are meticulously annotated with \emph{frame-level labels}, \emph{object boundaries}, and \emph{action sequences}. 
While these annotations are valuable, they are costly to produce and scale poorly with the number of modalities. 
Moreover, this approach assumes that models will be provided explicitly with complex handcrafted features to perform well, which contradicts the emerging capabilities of MLMs to autonomously extract and synthesize information from raw inputs. 
Inspired by recent findings~\cite{Plizzari2024outlookego}, M2AD adopts a minimal annotation approach, forcing the evaluated MLM to rely solely on the language and video understanding capabilities coming from their pre-training or fine-tuning to autonomously extract relevant information, reducing the reliance on costly, structured annotations.

Perspective diversity is another challenge in evaluating MLMs. 
In fact, perspective can be either egocentric (mimicking situations where it possible to have access to the same view of the user, like a VR), or exocentric (more similar to environments where one or more external cameras provide a view of the user appraoching the task to complete) and can introduce additional complexities. 
Egocentric views often suffer from occlusions, rapid scene changes, and limited visibility of fine details, while exocentric views may lack the contextual nuances of user interactions. 
Existing datasets typically focus on one perspective, failing to capture the full range of complexities encountered in real-world technical assistance scenarios. 
M2AD addresses this by including data recorded from both egocentric and exocentric perspectives, ensuring a more comprehensive evaluation of MLMs in diverse scenarios.

Finally, many benchmarks still allow models to rely on textual shortcuts rather than enforcing the integration of multi-modal inputs. 
This undermines the evaluation of how well models process data from diverse modalities, such as combining visual, textual, and contextual cues, to extract information. 
M2AD challenges models to integrate multiple modalities, ensuring a more robust assessment of their multi-modal reasoning capabilities.

% In summary, M2AD was developed to address the shortcomings of existing benchmarks by providing a more holistic, cost-effective, and realistic evaluation framework for MLLMs. By focusing on the complexities of technical assistance, adopting a minimal annotation approach, incorporating diverse perspectives, and challenging models to integrate multiple modalities, M2AD aims to drive progress in the field and bring us closer to achieving Artificial General Intelligence.

% TODO move here comments on the other datasets

\subsection{Annotation Process}
\label{sec:annotationprocess}

We designed the annotation process for the M2AD dataset to capture precise temporal intervals corresponding to each step in the assembly process of IKEA furniture. 
To ensure consistency and reliability, each video was manually segmented into annotated windows that aligned with the steps outlined in the instruction manuals. 
A key aspect of this process was determining the start of each step, which required a well-defined decision criteria. 
Specifically, the beginning of an interval was marked by the first frame in which either the components relevant to the step or the assembly action itself became visibly distinct. 
This approach aimed to minimize ambiguity and ensure uniformity across annotations.

\begin{figure}[htbp]
    \centering
    \includegraphics[width=.7\linewidth]{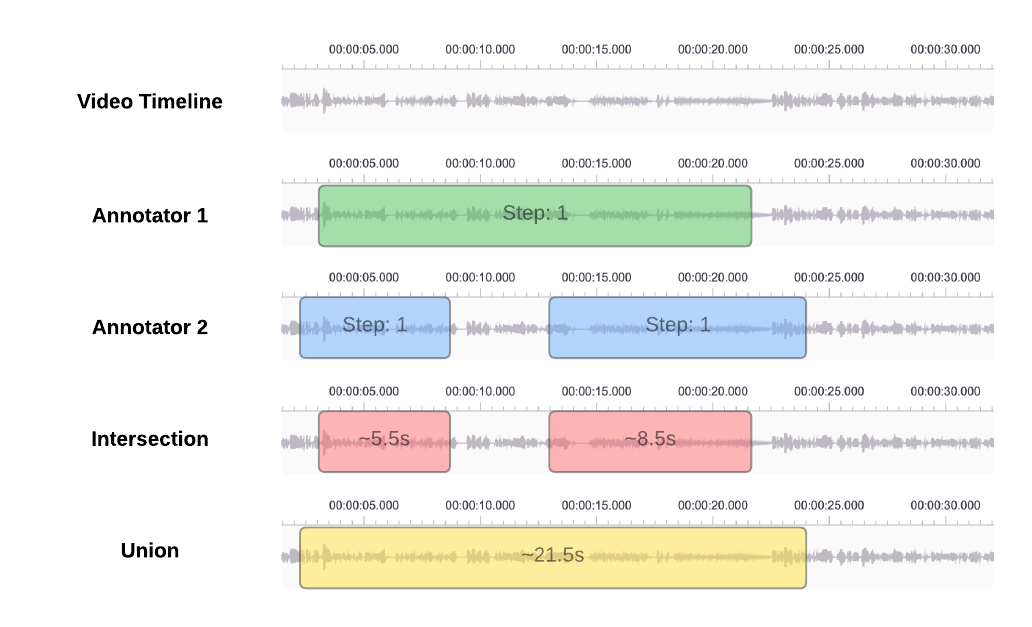}
    \caption{Example of an IoU computation for IAA.}
    \label{fig:iou_example}
\end{figure}

The inherent subjectivity of the annotation process posed a significant challenge.
To address and quantify this subjectivity we enrolled two independent annotators working on different overlapping subsets of the dataset. 
We quantified the subjectivity computing the Inter-Annotator Agreement (IAA), which we computed as the Intersection-over-Union (IoU) of the annotated intervals as illustrated in \Cref{fig:iou_example}. 
The annotators achieved an average of approximately $70\%$ IoU, reflecting a reasonable level of consistency given the complexities of the task.

During the annotation process, we encountered several additional challenges. 
These included handling duplicate step numbers, steps with multiple assembly versions, steps involving the assembly of multiple units, incomplete or interrupted steps, skipped steps, and variations in assembly sequences. 
Furthermore, frequent changes in the Point-of-View (POV) within the videos added another layer of complexity, requiring careful attention to ensure accurate and meaningful annotations. 
Despite these challenges, the process was designed to maintain a high standard of precision and reliability, ensuring the dataset's suitability for evaluating machine learning models in technical and assembly assistance tasks. 

\section{Experiments and Results}
\label{sec:experiments}

%TODO
In this section we outline the experiments we conducted on M2AD to establish baseline results on three tasks.

% In this section we outline the experiments we conducted on M2AD to establish baseline results on three tasks.
% Before going into the details of the experiments, we report the experimental settings.
% The tasks are \emph{step completion detection} (\Cref{sec:stepcompetiontracking}), \emph{step detection} (\Cref{sec:samestepdetection}) and \emph{step identification} (\Cref{sec:nextstepprediction}).
% In \Cref{sec:experimentscomment} we comment on the overall performances on the selected models across the three tasks.

\subsection{Settings}
\label{sec:settings}

In our experiments, we used the following MLMs: LLaVa-Video~\cite{zhang2024llava_video} 7B, LLaVa OneVision~\cite{li2024llava_onevision} 8B, Mantis-Idefics2~\cite{jiang2024mantis} 8B, MFuyu~\cite{jiang2024mantis} 8B, MolMo~\cite{deitke2024molmopixmoopenweights} 7B, Ovis~\cite{lu2024ovis} 3B, Qwen2-VL~\cite{wang2024qwen2vl} 8B and Pixtral~\cite{agrawal2024pixtral12b} 12B.
Models were loaded without quantization, when possible, and were prompted with zero-shot learning.
We conducted all experiments on a consumer level PC equipped with Intel Core i9 CPU, 64 GB RAM and a NVIDIA RTX 4090 GPU.

We selected all open models for reproducibility.
Moreover, we conducted the experiments on consumer level hardware both for easier reproducibility and to better simulate a real-world scenario.
In fact, as often happens in industrial settings where technical assistance may be offered, it is not always the case that (1) data can be processed on external servers or by third party APIs and (2) the available resources allow to run models of any size.

% ed to solve a binary classification problem, generating \texttt{`0'} or \texttt{`1'} depending on the predicted class, while in the third we modelled the problem as a multi-class classification task where the model had to predict the ...
% ng textual instructions, observed frames (often with downsampled frame-rate to fit context window) and manual page or pages using zero-shot learning.
% classification metrics (accuracy, precision, recall and F1).

\subsection{Step Completion Detection}
\label{sec:stepcompetiontracking}

\begin{table}[!ht]
\begin{center}
% \resizebox{0.45\textwidth}{!}{
\begin{tabular}{lS[table-format=2.2]S[table-format=2.2]S[table-format=2.2]S[table-format=2.2]}

    \toprule

    \multirow{2}{*}{\textbf{Model}} & \multicolumn{4}{c}{\textbf{Metric [\%]}} \\ \cmidrule(l){2-5}
    & {\textbf{Accuracy}} & {\textbf{Precision}} & {\textbf{Recall}} & {\textbf{F1}} \\

     \midrule
     
     \textbf{LLaVa-Video} & $\mathbf{56.79}$ & 59.13 & 56.79 & 53.84 \\
     \textbf{LLaVa-OneVision} & 53.42 & 57.16 & 53.42 & 46.41 \\
     \textbf{MFuyu} & 50.04 & 50.62 & 50.04 & 34.81 \\
     \textbf{MolMo} & 50.12 & 56.84 & 50.12 & 33.88 \\
     \textbf{Ovis} & 55.65 & 55.67 & 56.65 & 55.62 \\
     \textbf{Qwen2-VL} & \underline{56.71} & 58.92 & 56.71 & 53.86 \\
     \textbf{Pixtral} & 50.85 & 55.04 & 50.85 & 37.98 \\
     \textbf{Mantis-Idefics2} & 49.71 & 49.58 & 49.71 & 45.44 \\
     
    \bottomrule

\end{tabular}  
% }
\end{center}
\caption{Step completion detection results (\textbf{best score}, \underline{second-best score}).}
\label{tab:experiment_1_results}
\end{table}

This experiment assesses whether an MLM can determine, given the current assembly state (passed as frames from the video) and a relevant page from the instruction manual, whether the step on the page has been successfully completed by the user.
The model, instructed to act as a technical assistant, is presented with a sequence of video frames and the associated page from the instruction manual.
In this experiment, each model was prompted to determine whether the assembly step visualised in the clip could be considered completed with respect to the step indicated in the manual page.
We selected frames from the end and the start of each assembly segment as positive and negative samples.

We report the results of the step completion tracking experiment in \Cref{tab:experiment_1_results}.
Looking at the accuracy scores, besides the LLava-Video \cite{zhang2024llava_video}, Qwen2-VL \cite{wang2024qwen2vl} and Ovis \cite{lu2024ovis} models, all other models return a score close to that of a random classifier.
Interestingly, the Ovis \cite{lu2024ovis} model achieves an accuracy score comparable to models with more than double the number of parameters.

\subsection{Step Detection}
\label{sec:samestepdetection}

\begin{table}[!ht]
\begin{center}
% \resizebox{0.45\textwidth}{!}{
\begin{tabular}{lS[table-format=2.2]S[table-format=2.2]S[table-format=2.2]S[table-format=2.2]}

    \toprule

    \multirow{2}{*}{\textbf{Model}} & \multicolumn{4}{c}{\textbf{Metric [\%]}} \\ \cmidrule(l){2-5}
    & {\textbf{Accuracy}} & {\textbf{Precision}} & {\textbf{Recall}} & {\textbf{F1}} \\
    
    \midrule
    
    \textbf{LLaVa-Video} & $\mathbf{59.16}$ & 59.34 & 59.16 & 58.95 \\
    \textbf{LLaVa-OneVision} & 50.00 & 50.00 & 50.00 & 35.12 \\
    \textbf{MFuyu} & 50.40 & 51.25 & 50.40 & 40.31 \\
    \textbf{MolMo} & {50.00} & {25.00} & {50.00} & {33.33} \\
    \textbf{Ovis} & 50.04 & 58.34 & 50.04 & 33.49 \\
    \textbf{Qwen2-VL} & \underline{55.90} & 58.68 & 55.90 & 52.06 \\
    \textbf{Pixtral} & 54.72 & 54.87 & 54.72 & 54.35 \\
    \textbf{Mantis-Idefics2} & 50.00 & 50.00 & 50.00 & 33.47 \\ 
    
    \bottomrule

\end{tabular}  
% }
\end{center}
\caption{Step detection results (\textbf{best score}, \underline{second-best score}).}
\label{tab:experiment_2_results}
\end{table}

This experiment assesses whether an MLM can determine, given the current assembly state (passed as frames from the video) and one or more pages from the instruction manual, whether the ongoing step is present in one of the pages.
The model, instructed to act as a technical assistant, is presented with a sequence of video frames along with a selection of two instruction manual pages, including both relevant and distractor pages. 
In this experiment, each model was prompted to determine whether the pages contained instructions relevant to the step visualised in the frames. 
We selected frames from the start of each assembly step and two instruction manual pages: The pages with the correct step as positive samples and 
the pages adjacent to the correct one, therefore not containing that assembly step, as negative samples.

We report the results of the step detection experiment in \Cref{tab:experiment_2_results}. 
In this case only LLaVa-Video \cite{zhang2024llava_video} and Qwen2-VL \cite{wang2024qwen2vl} returned results sufficiently better than a random baseline, with LLaVa-Video being again the best performing model.

\subsection{Step Identification}
\label{sec:nextstepprediction}

\begin{table}[!ht]
\begin{center}
% \resizebox{0.45\textwidth}{!}{
\begin{tabular}{lS[table-format=2.2]S[table-format=2.2]S[table-format=2.2]S[table-format=2.2]}

    \toprule

    \multirow{2}{*}{\textbf{Model}} & \multicolumn{4}{c}{\textbf{Metric [\%]}} \\ \cmidrule(l){2-5}
    & {\textbf{Accuracy}} & {\textbf{Precision}} & {\textbf{Recall}} & {\textbf{F1}} \\
    
    \midrule
    
    \textbf{LLaVa-Video} & \underline{23.77} & 73.64 & 13.50 & 19.03 \\
    \textbf{LLaVa OneVision} & 20.35 & 22.64 & 31.26 & 23.94 \\
    \textbf{MFuyu} & 21.82 & 38.66 & 29.40 & 30.04 \\
    \textbf{MolMo} & $\mathbf{79.56}$ & 86.53 & 85.78 & 86.03 \\
    \textbf{Ovis} & 14.49 & 38.18 & 9.02 & 12.82 \\
    \textbf{Qwen2-VL} & 17.18 & 63.40 & 9.83 & 13.73 \\
    \textbf{Pixtral} & 20.03 & 40.90 & 15.46 & 20.37 \\
    \textbf{Mantis-Idefics2} & 18.24 & 27.63 & 25.04 & 24.27 \\

    \bottomrule

\end{tabular}  
% }
\end{center}
\caption{Step identification results (\textbf{best score}, \underline{second-best score}).}
\label{tab:experiment_3_results}
\end{table}

%This experiment assesses whether an MLM can determine, given the current assembly state (passed as frames from the video) and one or more pages from the instruction manual, which on the steps in the pages is the ongoing one.
This experiment assesses whether an MLM can determine, given the current assembly state (passed as frames from the video) and one or more pages from the instruction manual, the number of the assembly step currently being executed.
The model, instructed to act as a technical assistant, is presented with a sequence of video frames along with two pages of the corresponding instruction manual, one of which contains the current step, while the other serves as a distractor.
In this experiment, the model was prompted to respond with the exact assembly step being executed in the frames from the clip. 
We selected both the beginning and end frames from each assembly step and a pair of consecutive pages including the one with the correct step and the next one in the assembly procedure.
Differently from the previous experiments, this is now a multi-class classification problem, therefore, each sample's label is the associated assembly step number.
On average, this meant presenting the model with \textbf{four} alternative steps shown on the two pages from the instruction manual, where only one was the correct one.
To ensure the results are interpretable, we propose two random-guessing baselines depending on whether we assume the models to exploit Optical Character Recognition (OCR) capabilities or not: (1) picking one step at random among the set of all possible step numbers in our dataset: $\nicefrac{1}{59}$ and (2) picking one step at random among the four steps (on average) presented in the pages from the instruction manual: $\nicefrac{1}{4}$.

We report the results of the step detection experiment in \Cref{tab:experiment_3_results}. 
In this case almost every model struggles with this task, except for MolMo \cite{deitke2024molmopixmoopenweights} which achieved good result despite its lack of support for interleaved text-image reasoning. Specifically, this means that the MolMo model was prompted on concatenated images of frames and instruction manual pages one next to the other, and based its performance on spatial reasoning (using references to "left" and "right" sides of the image). Furthermore, the model's unique ability of "pointing" to entities or objects in an image \cite{deitke2024molmopixmoopenweights} is likely to have contributed to its overall impressive performance.

\subsection{Comment}
\label{sec:experimentscomment}

%The experiments conducted on the M2AD dataset highlighted the critical role of instruction-following fine-tuning in zero-shot settings
%Some models, such as MolMo and Ovis, failed to respond adequately to the evaluation prompts, resulting in null evaluation on step detection and identification.
%We hypothesise that models not explicitly trained or fine-tuned for instruction-following struggle to perform effectively in such scenarios because they fail at interpreting instructions.
Although the experiments in this study were conducted within a resource-constrained setting, which inherently limited the amount of information available to the models, we believe that the results still offer valuable insight.

Firstly, the experiments reveal the importance of multi-image reasoning, interleaved text-image reasoning and specifically spatial reasoning in images. 
These capabilities are crucial for guiding users through complex tasks, where understanding visual sequences and integrating them with textual instructions is essential.
In fact, in contexts which involve user interaction, step-by-step guidance almost always relies on the interplay between visual and textual cues.

Furthermore, due to the real-time application requirement for the use-case scenario of domain-specific assistance (such as furniture assembly), achieving good performance while working with data as minimal in size as possible is critical to the usability of any model. This is important both to having models that can run on consumer-level hardware to avoid privacy issues and, more importantly, to ensure user-model interactions with minimal latency.

However, at the same time, the results underscore the impact of architectural and hardware limitations on model performance.
Constraints such as processing limitations for large context windows and memory limits for multi-frame analysis significantly restrict the ability of MLMs to process and interpret complex scenarios. 
These limitations mean that models often operate with incomplete or truncated data, hindering their ability to fully grasp the intricacies of multi-modal tasks.

\section{Conclusion}
\label{sec:conclusion}

In this paper, we introduced M2AD, a novel dataset designed to evaluate MLMs in the context of technical assistance tasks, specifically in furniture assembly. 
Through three carefully designed experiments, we evaluated the capabilities of openly available MLMs running on consumer-level hardware in (1) tracking step completion, (2) comparing steps from instruction manual pages and video frames to determine their correspondence, and (3) detecting the current assembly step number being executed.
Our findings highlight that some MLMs exhibit a good ability to handle complex tasks with minimal annotations, underscoring their potential to reduce the reliance on highly detailed or fine-grained annotation efforts, which characterises other datasets, suggesting a promising direction for developing cost-effective annotation strategies in multimodal settings.
However, the results also revealed notable limitations, particularly about the MLMs' capabilities of reasoning on images, comparing images as fundamentally different in nature as instruction manual pages with real-life images of furniture assembly procedures.
In general, the main limitations emerge as a combination of architectural and hardware limitations, which forced us to heavily restrict the amount of information the MLMs could work with, both in terms of quantity and resolution.
On the other hand, MolMo \cite{deitke2024molmopixmoopenweights} and LLaVa-Video \cite{zhang2024llava_video} clearly show two important and promising traits for the future of MLMs design and development: The capability of pointing at entities/objects in an image, and the capability of working with multiple images interleaved with text, which replicate exactly two important traits in the domain of technical assistance.
%evalauted models instruction-following capabilities. 
%In fact, while instruction-tuned MLMs demonstrated consistent performance in differentiating relevant from irrelevant manual pages and predicting the current step, they still struggled with tasks involving closely related procedural steps. 
%This indicates a need for further refinement in model architectures and training methodologies to improve reliability in such scenarios.

Building on the insights emerged from our experiments, we can suggest multiple future research directions worth exploring. 
One promising direction is to look deeper into pre-training or fine-tuning approaches for multimodal models to understand and improve the alignment between modalities and the instruction following aspect, which resulted limited in many of the currently available open-access models. 
Additionally, we are interested in exploring the dependency from annotations of MLMs across several tasks, to understand up to which point it is possible to rely on model understanding capabilities only to parse a given situation starting only from the raw input data to assess the stage of completion of a task and predict the next steps.
Finally, we are interested in extending the evaluation of MLMs as assistants to a broader range of technical and domain-specific settings, including other domains such as vehicles repair, healthcare and programming as they could provide a better understanding of actual MLMs cross-domain generalisation in assisting humans.

\begin{comment}
We introduce M2AD - A new dataset for the evaluation of Multimodal LLMs in the setting of technical assistance tasks, more specifically of furniture assembly.
We presented three experiments for the evaluation of MLLMs which evaluated the capabilities in (1)tracking a step's completion, (2)confronting steps from instruction manual pages and frames, detecting whether they are the same or not, and (3)predicting which step exactly is currently being executed given frames and pages from the instruction manual.

The results presented in the previous chapter show that some models, particularly those with instruction tuning skills demonstrate the ability to handle complex tasks with input with minimal annotations, supporting the potential for MLLMs to reduce the need for highly detailed or fine-grained annotations, and sugesting a feasible pathway for cost-effective annotation strategies.
Furthermore, while the results show that certain MLLMs are capable of detecting step transitions, they reveal the limitations of models with less preparation for instruction following. Overall, models with instruction-tuning performed consistently well, showing their ability to differentiate between relevant and irrelevant manual pages to some extent, and in predicting the current step being executed, though critical improvements are needed to handle tasks with closely related procedural steps reliably.
\end{comment}

\begin{credits}
\subsubsection{Material}
\url{https://github.com/ftoschi14/M2AD-From-Instructions-to-Assistance}

\subsubsection{\discintname}
The authors have no competing interests to declare.
% The authors have no competing interests to declare that are relevant to the content of this article.
\end{credits}
%
% ---- Bibliography ----
%
% BibTeX users should specify bibliography style 'splncs04'.
% References will then be sorted and formatted in the correct style.
%
% \bibliographystyle{splncs04}
% \bibliography{mybibliography}
%% Note that this preceding line implies that you store your BibTeX references in a file called 'mybibliography.bib'. If you instead store your references in a file with a different name, for instance 'references.bib', the preceding line should read '\bibliography{references}'. Whatever you do, DO NOT put the file name extension .bib inside the \bibliography command; this will trip up LaTeX compilers. 
%
% If you do not want to use BibTeX, you can also type up the bibliography exactly as you see fit, using the following structure:

\bibliographystyle{splncs04}
% \bibliography{bibliography}
\bibliography{minimal_bibliography}

\end{document}